\documentclass{isprs} 
\usepackage{subfigure}
\usepackage{setspace}
\usepackage{geometry} 
\usepackage{epstopdf}
\usepackage[labelsep=period]{caption}  
\usepackage[british]{babel} 
\usepackage[hang]{footmisc}
\usepackage{float} 
\usepackage{tabularx}
 \usepackage{booktabs}
 \usepackage{natbib}
 \usepackage{siunitx}
\usepackage{amsmath}
\usepackage{xcolor}
\usepackage{float}
\usepackage{float}
\usepackage{tabularx,booktabs}
\usepackage{siunitx}
\usepackage{subfigure} 
\usepackage{url}
\begin{document}
\title{3D Density-Gradient based edge detection on Neural Radiance Fields (NeRFs) for geometric reconstruction}
\date{}

\author{Miriam Jäger\thanks{Corresponding author}, Boris Jutzi}
\address{
	Institute of Photogrammetry and Remote Sensing, Karlsruhe Institute of Technology, Germany \\(miriam.jaeger, boris.jutzi)@kit.edu
}

\abstract{Generating geometric 3D reconstructions from Neural Radiance Fields (NeRFs) is of great interest. However, accurate and complete reconstructions based on the density values are challenging. The network output depends on input data, NeRF network configuration and hyperparameter. As a result, the direct usage of density values, e.g. via filtering with global density thresholds, usually requires empirical investigations.
Under the assumption that the density increases from non-object to object area, the utilization of density gradients from relative values is evident. As the density represents a position-dependent parameter it can be handled anisotropically, therefore processing of the voxelized 3D density field is justified.
In this regard, we address geometric 3D reconstructions based on density gradients, whereas the gradients result from 3D edge detection filters of the first and second derivatives, namely Sobel, Canny and Laplacian of Gaussian. The gradients rely on relative neighboring density values in all directions, thus are independent from absolute magnitudes. Consequently, gradient filters are able to extract edges along a wide density range, almost independent from assumptions and empirical investigations.
Our approach demonstrates the capability to achieve geometric 3D reconstructions with high geometric accuracy on object surfaces and remarkable object completeness. 
Notably, Canny filter effectively eliminates gaps, delivers a uniform point density, and strikes a favorable balance between correctness and completeness across the scenes.}



\maketitle



\section{INTRODUCTION}

Neural Radiance Fields (NeRFs) \citep{mildenhall_et_al_2020} pioneered computer graphics and computer vision by enabling the rendering of novel views through view synthesis from neural networks. These networks estimate density and color values for each position in 3D space based on input image data and camera poses. Generating accurate and complete 3D reconstructions from Neural Radiance Fields (NeRFs) is of interest in the field of photogrammetry. Through the utilization of estimated density values, NeRF based 3D reconstructions are possible. More precisely, by considering the density as a kind of pseudo-probability for the occurrence of an object in 3D space \citep{jaeger2023hololens}. 

Nevertheless, the filtering with global density thresholds is empirical and requires sufficient analysis of its geometric correctness. Accordingly, the 3D reconstruction depends on the chosen density threshold and often yields noisy and incomplete surfaces \citep{neuralangelo,NeuS}. The assumption that the density increases from non-object to object area, motivates the processing of the 3D scene in terms of its density gradients. As the density represents a position-dependent parameter, it can be addressed anisotropically, justifying a ray-independent sampling. For this reason, we propose to perform geometric 3D reconstruction with respect to density gradients, while the key aspect is the utilization of 3D gradient filter for 3D edge detection. This allows the extraction of edges along a wide density range based on gradients of relative neighboring values.

We introduce a straightforward workflow for enabling geometric 3D reconstruction from NeRFs with the 3D density gradients based on the first and second derivative, while using gradient filter for edge detection in the voxelized 3D density field. In order to evaluate the geometric accuracy and robustness of our framework, we address the DTU benchmark dataset \citep{dtu} with different types of real objects, which feature different sizes, structures, materials, textures and colors.
\begin{figure}[H]
\begin{center}
		\includegraphics[width=0.69\columnwidth]{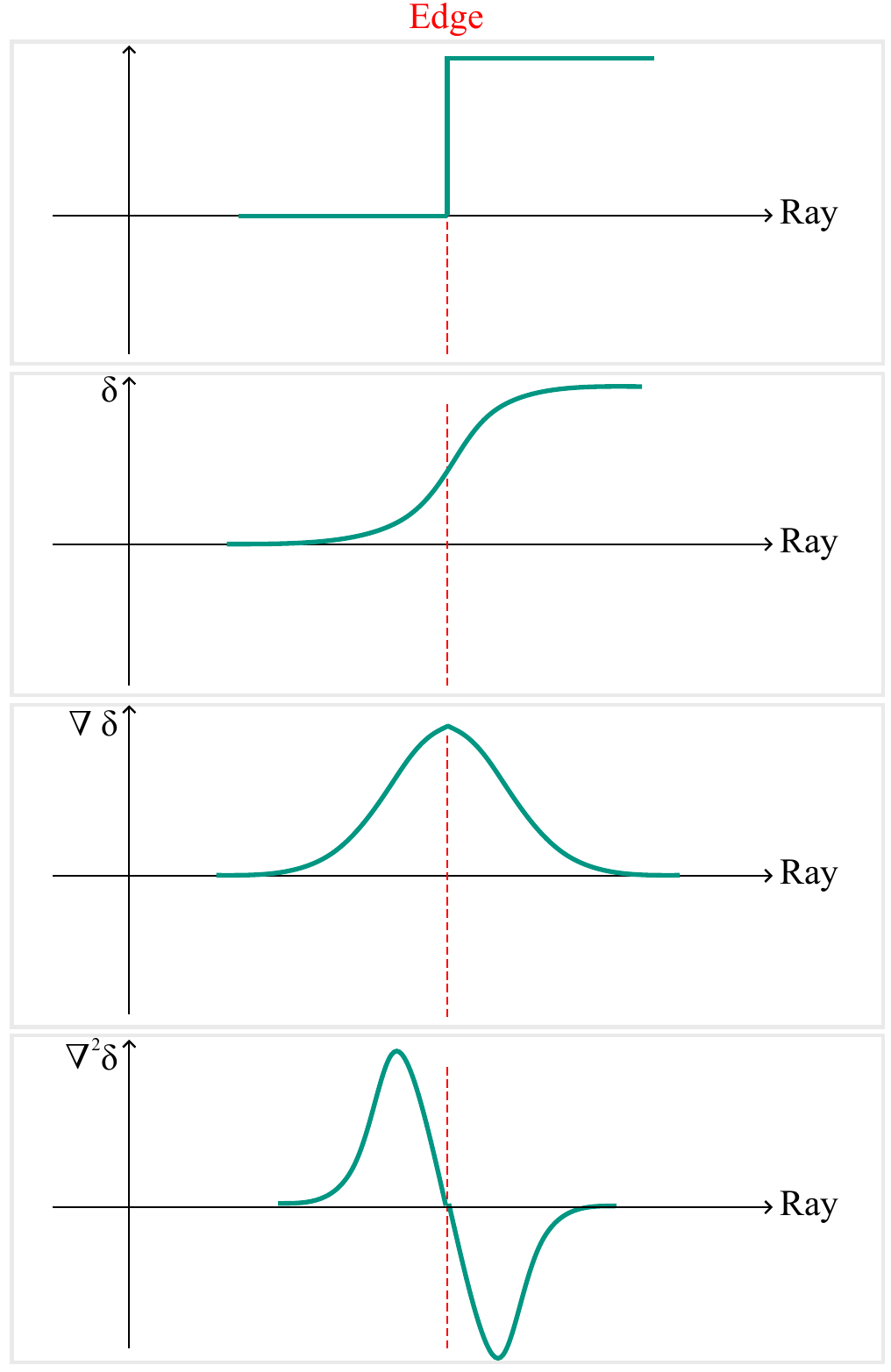}
	\vspace{-3mm}
	\caption{Density Gradient e.g. for 1D on a ray. The illustrations display the characteristics of the density values exemplarily as they would occur during ray tracing into the direction of an object. From top to bottom: An ideal edge in a binary non-object to object space, the raw density values, the first derivative of the density values (edge at the maximum), the second derivative of the density values (edge at the zero crossing).}
\label{fig:Density_Gradient}
\end{center}
\end{figure}

\section{RELATED WORK}\label{RELATED WORK}
In this section, we briefly summarize related work to our research. Firstly, we give an overview on basic, recent research and developments on NeRFs. Following this, we address recent research on neural surface reconstructions.

\paragraph{Neural Radiance Fields}\label{sRW_Nerfs}

The foundation for Neural Radiance Fields (NeRFs) was established by Scene Representation Networks (SRNs) \citep{SceneRepresenationNetworks}. Their underlying principle is modeling the scene as a function of 3D coordinates within it. It was followed by the groundbreaking research work of Neural Radiance Fields \citep{mildenhall_et_al_2020}. The network enables the estimation of color and density values for each 3D position through 6D camera poses and associated 2D images by training a neural network with multi-layer perceptrons (MLPs). 

The vanilla NeRF was followed by thousands of publications driving research and development in various domains.
Scalability enhancements are demonstrated by Mega-NeRF \citep{Mega-nerf} and Block-NeRF \citep{Block-nerf}, which employ data partitioning and the training of several NeRFs.
Bundle Adjusting Radiance Fields (BaRF) \citep{Barf} and Gaussian Activated Radiance Fields (GaRF) \citep{Garf} address the task of a camera pose estimation. Dynamic contributions use time as an additional input dimension for time-dependent rendering \citep{D_nerf} or for preventing the occurrence of artifacts due to dynamic pixels \citep{dynamic_nerf}. Several Methods such as AdaNeRF \citep{adanerf}, FastNeRF \citep{fastnerf} and Instant NGP \citep{mueller_et_al_2022} focus on faster training or rendering. While Instant NGP uses a combination of small MLPs and spatial hash table encoding. Neuralangelo \citep{neuralangelo} adapts Instant NGP and combines hash grids with neural surface rendering for high-fidelity surface reconstruction. Besides the neural methods, non-neural research like Plenoxels \citep{plenoxels} have been introduced.

\paragraph{Neural Surface Reconstructions}\label{sRW_Nerfs}
Regarding neural surface reconstructions Unisurf
\citep{unisurf} learns implicit surfaces, by addressing the occupancy along rays. Several works such as NeuS \citep{NeuS} and VolSDF \citep{volsdf} represent the scene by neural Signed Distance Functions (SDFs) \citep{deepsdf}. Neuralwarp \citep{NeuralWarp} builds on VolSDF, whereas using Structure from Motion information to guide surface optimizations. 


\section{METHODOLOGY}\label{sec:METHOD}
Firstly, in Section \ref{sec:gradient} describes the principal motivation for density gradients underlying our framework. Secondly, in Section \ref{sec:First_derivative} and Section \ref{sec:Second_derivative} the first and second derivative calculation for density gradients is explained. Finally, Section \ref{sec:evaluation} outlines the evaluation process, which focuses on completeness and correctness.

\subsection{Density Gradient}\label{sec:gradient}
Reconstructions based on filtering the NeRFs density output by global density thresholds requires adaptive adjustments, since the density values behavior differ for various NeRFs, datasets, hyperparameters and network configurations. Accordingly, the 3D reconstruction depends on the chosen threshold and does not provide optimal, noisy or incomplete reconstructions \citep{NeuS,unisurf,neuralangelo}.

Several previous works consider ray-based 3D reconstruction with NeRFs or SDFs \citep{unisurf,NeuS,NeuralWarp}.
Nevertheless, the density values in principle are anisotropic and position-dependent. For this reason, we propose to process the geometric 3D reconstruction in the dense voxelized 3D density field.
With the aim of performing position-dependent 3D reconstructions, regardless of global density thresholds, we introduce 3D gradient filters. 
To identify edges characterized by variations in magnitudes, hence density values, we extend from two to the three-dimensional edge filter among the 3D density field.
In doing so, the density gradients instead of the raw density values from NeRFs are regarded, since the density value increases towards the object. The extraction of the edges can rely on the first as well as the second derivative of the density, see Figure \ref{fig:Density_Gradient}. Thereby, we guarantee anisotropy as well as the consideration of neighborhoods in the reconstruction process.

\subsection{First Derivative}\label{sec:First_derivative}
Edges in images as well as in 3D voxel space can be detected based on the first derivatives, i.e. the corresponding density gradients in this case.

\paragraph{Sobel filter}\label{sec:sobel}
We address the well-established Sobel filter \citep{sobel} for edge detection, which performs a smoothing orthogonal to the first derivative. As the processing is done in the 3D density field, the 3D Sobel filter is built up of the following components for each direction x, y and z, e.g., for the x-direction for the central element \citep{sobel}:
\begin{equation}
  \begin{aligned}
	s_{\text{-}1} \text{=} 
     \begin{bmatrix}
     	   \text{-}1&\text{-}2 &\text{-}1 \\
          \text{-}2&\text{-}4 &\text{-}2 \\ 
          \text{-}1&\text{-}2 &\text{-}1
     \end{bmatrix},
	s_{0} \text{=} 
     \begin{bmatrix}
     	   0&0 &0 \\
          0&0 &0 \\ 
          0&0 &0
     \end{bmatrix},
	s_{1} \text{=} 
     \begin{bmatrix}
     	   1&2 &1 \\
          2&4 &2 \\ 
          1&2 &1
     \end{bmatrix},
     \end{aligned}     
\end{equation}
and delivers the density gradients $\text{G}_{\delta,x}$, $\text{G}_{\delta,y}$ and $\text{G}_{\delta,z}$ in direction x, y and z in the density field.
The total Sobel gradient $\Delta_{\delta,\text{Sobel}}$ for each sample in the density field is further given by
\begin{equation}\label{equ:1}
    \begin{aligned}
    \Delta_{\delta,\text{Sobel}} &= \sqrt{\text{G}_{\delta,x}^2 + \text{G}_{\delta,y}^2 + \text{G}_{\delta,z}^2}.
    \end{aligned}
\end{equation}

\paragraph{Canny filter}\label{sec:canny}
Furthermore, we address Canny filter \citep{canny} for 3D edge detection, while it offers an improved edge detection in contrast to the Sobel filter.
The gradient calculation based on the density value, such as described for Sobel filter, is first preceded by a Gaussian smoothing in the 3D density field to suppress noise. This is followed by gradient magnitude thresholding with a lower and upper relative threshold on the density gradients for edge detection. Finally, a hysteresis method is used to track strong edges and suppress weak ones at the same time. The final density gradient values based on Canny filter are referred as $\Delta_{\delta,\text{Canny}}$ in the following.
From this method, we expect to extract a wide variation of edges in the density field and, in particularly, detect the object edges through the final step of hysteresis.

\subsection{Second Derivative}\label{sec:Second_derivative}
Edge detection in images and 3D voxel space can not only be performed based on the first derivative, as it is the case with the Sobel filter and the extension of the Canny filter. The second derivative provides a basic approach to edge detection based on differences of neighboring values, while edges result from the zero crossings.
Since the second derivative is usually sensitive to noises, a previous smoothing of the values is essential. 
\paragraph{Laplacian of Gaussian} From this point Laplacian of Gaussian filter \citep{log} (LOG), also referred as Marr-Hildreth operator, is suitable. It combines the second derivative with a Gaussian filter in order to smooth the values. For fast implementation the Difference of Gaussians (DoG) can be applied, which approximates the LOG. Similar to the filter of the first derivative, we apply the filter on the voxelized 3D density field and refer it as $\Delta^2_{\delta,\text{LOG}}$ in the following.
\subsection{Evaluation}\label{sec:evaluation}

\paragraph{Completeness}
In general, we report qualitative completeness on the basis of the resulting 3D reconstructions. Furthermore, the completeness is measured quantitatively. The reconstructions from voxelized 3D density field include predicted points inside the object and the reference point cloud contains large gaps.
We report the number of points and percentages covered by the NeRF reconstructions within a distance threshold of a maximum distance from reference. A higher score indicates higher object completeness.

\paragraph{Correctness}
To evaluate the geometric accuracy of the 3D reconstructions quantitative as well as qualitative, Chamfer cloud-to-cloud distance is applied from the DTU dataset evaluation script \citep{dtu}. 
We report both the distance from data to reference (data-to-reference) and vice versa (reference-to-data). While the reference point cloud has gaps, the data to reference distance as well as the reference to data distance are interpreted as accuracy or correctness.



\begin{figure*}[h!!]
	\centering
\hspace{0.5cm}
\raisebox{\dimexpr 0cm-\height}{reference}
\hspace{2.0cm}
\raisebox{\dimexpr 0cm-\height}{$\delta_{\text{t=50}}$}
\hspace{2.4cm}
\raisebox{\dimexpr 0cm-\height}{$\Delta_{\delta,\text{Sobel}}$}
\hspace{2.4cm}
\raisebox{\dimexpr 0cm-\height}{$\Delta_{\delta,\text{Canny}}$}
\hspace{2.0cm}
\raisebox{\dimexpr 0cm-\height}{$\Delta^2_{\delta,\text{LOG}}$}\hspace{4.0cm}\\
\rotatebox{90}{$\,\,\,\,\,\,\,\,\,\,\,\,\,\,\,\,$scan24}
\hspace{0.3cm}
\subfigure{\label{fig:Ficus_HoloLens_intern}
	\includegraphics[width=0.145\linewidth]{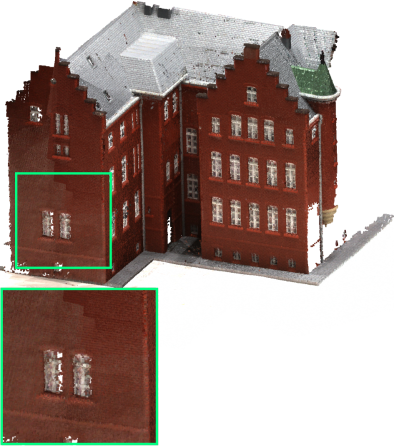}}\hspace{0.6cm}
\subfigure{\label{fig:Ficus_HoloLens_intern_pose}  
     \includegraphics[width=0.145\linewidth]{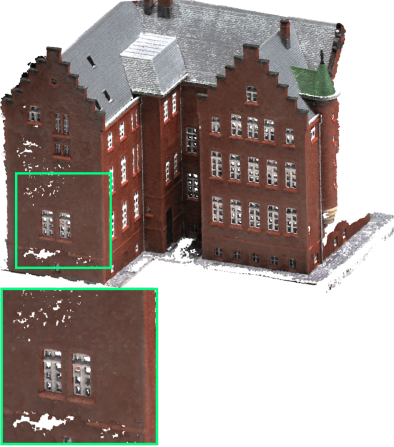}} \hspace{0.6cm}
\subfigure{\label{fig:Ficus_HoloLens_COLMAP_pose}  
     \includegraphics[width=0.145\linewidth]{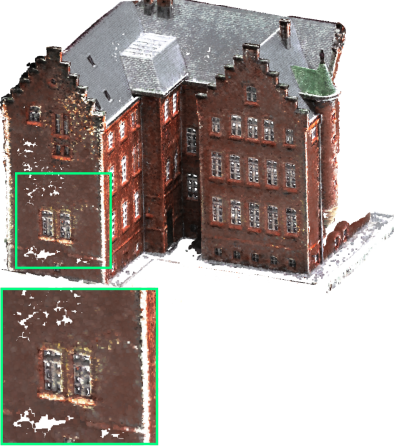}}\hspace{0.6cm}
\subfigure{\label{fig:Ficus_HoloLens_COLMAP_MVS}
	\includegraphics[width=0.145\linewidth]{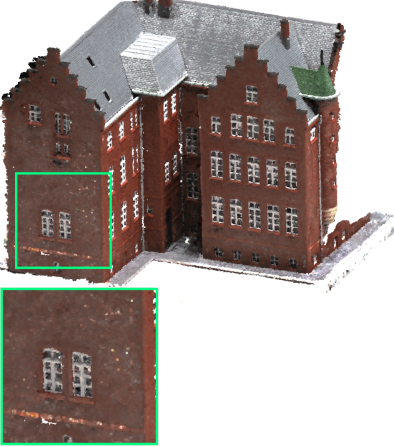}}\hspace{0.6cm}
\subfigure{\label{fig:Ficus_HoloLens_COLMAP_MVS}
	\includegraphics[width=0.145\linewidth]{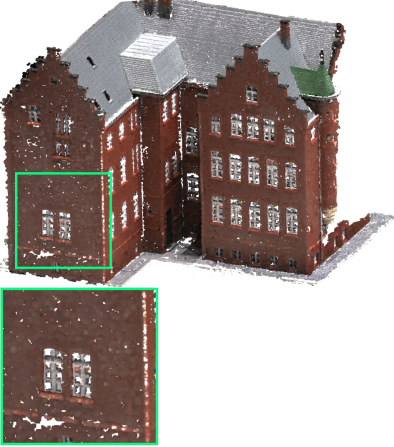}}\\
\rotatebox{90}{$\,\,\,\,\,\,\,\,\,\,\,\,\,\,\,\,$scan37}
\hspace{0.2cm}
\subfigure{\label{fig:Ficus_HoloLens_intern}
	\includegraphics[width=0.145\linewidth]{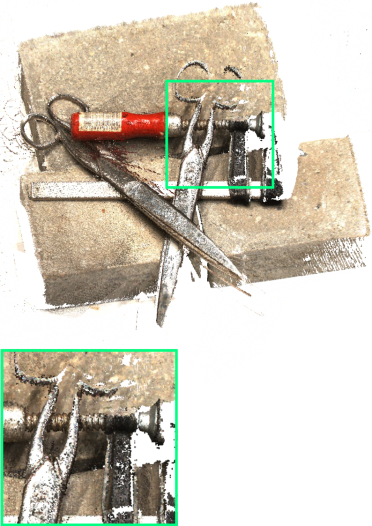}} \hspace{0.57cm}
\subfigure{\label{fig:Ficus_HoloLens_intern_pose}  
     \includegraphics[width=0.145\linewidth]{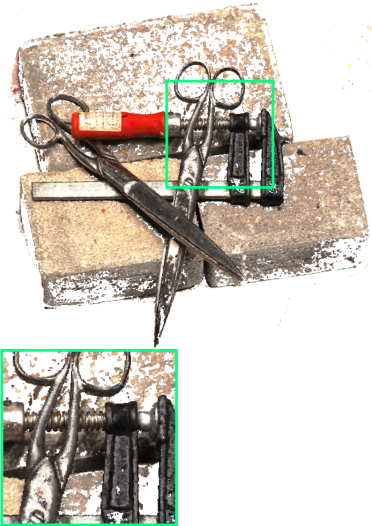}} \hspace{0.57cm}
\subfigure{\label{fig:Ficus_HoloLens_COLMAP_pose}  
     \includegraphics[width=0.145\linewidth]{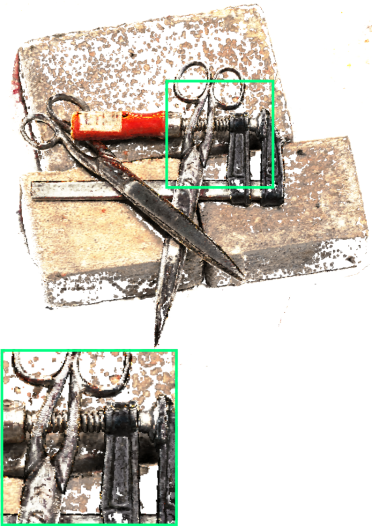}}\hspace{0.57cm}
\subfigure{\label{fig:Ficus_HoloLens_COLMAP_MVS}
	\includegraphics[width=0.145\linewidth]{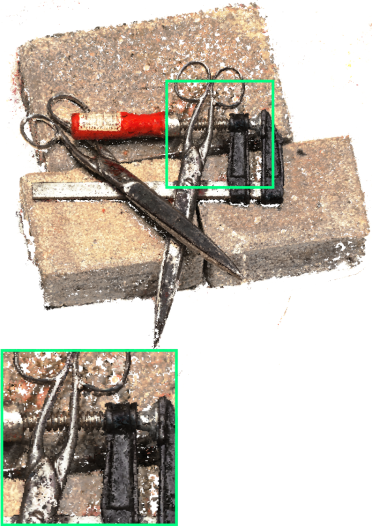}}\hspace{0.57cm}
\subfigure{\label{fig:Ficus_HoloLens_COLMAP_MVS}
	\includegraphics[width=0.145\linewidth]{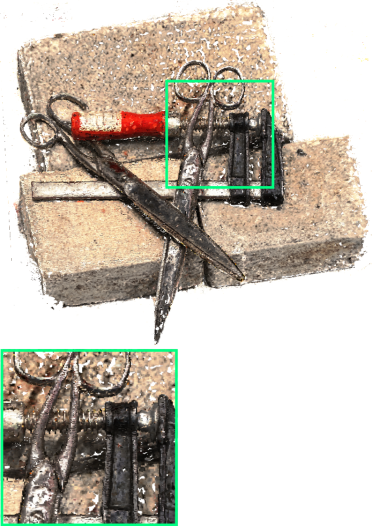}}\\
\rotatebox{90}{$\,\,\,\,\,\,\,\,\,\,\,\,\,\,\,\,$scan40}
\hspace{0.3cm}
\subfigure{\label{fig:Ficus_HoloLens_intern}
	\includegraphics[width=0.145\linewidth]{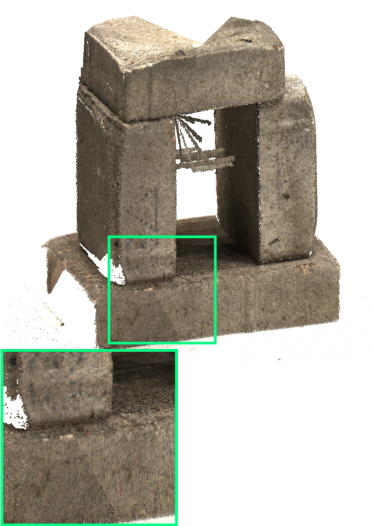}}  \hspace{0.5cm}
\subfigure{\label{fig:Ficus_HoloLens_intern_pose}  
     \includegraphics[width=0.145\linewidth]{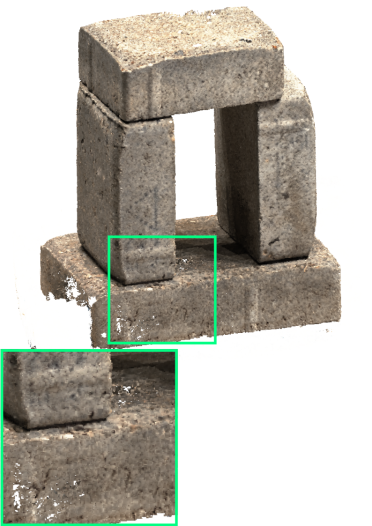}} \hspace{0.5cm}
\subfigure{\label{fig:Ficus_HoloLens_COLMAP_pose}  
     \includegraphics[width=0.145\linewidth]{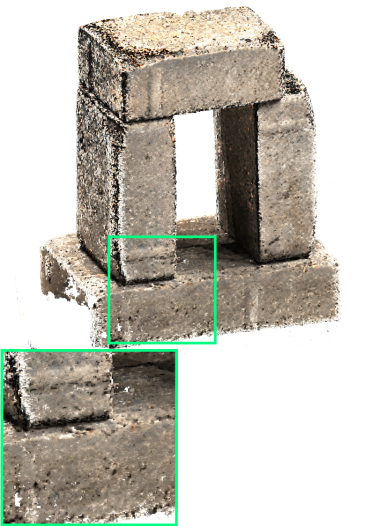}} \hspace{0.5cm}
\subfigure{\label{fig:Ficus_HoloLens_COLMAP_MVS}
	\includegraphics[width=0.145\linewidth]{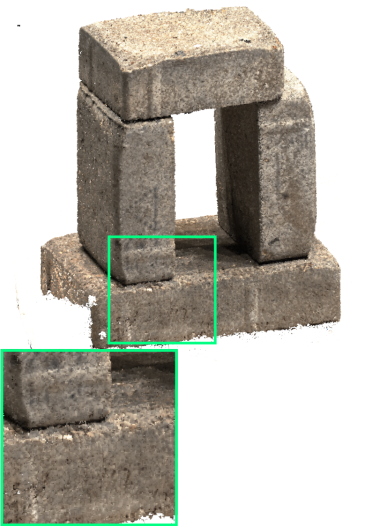}} \hspace{0.5cm}
\subfigure{\label{fig:Ficus_HoloLens_COLMAP_MVS}
	\includegraphics[width=0.145\linewidth]{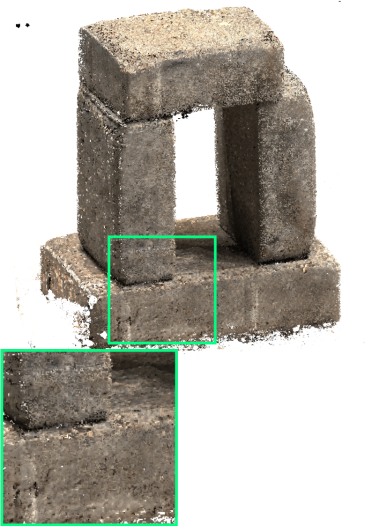}}\\
\rotatebox{90}{$\,\,\,\,\,\,\,\,\,\,\,\,\,\,\,\,$scan55}
\hspace{0.2cm}
\subfigure{\label{fig:Ficus_HoloLens_intern}
	\includegraphics[width=0.145\linewidth]{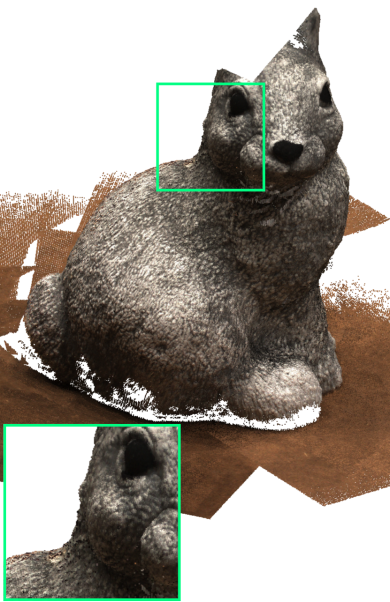}} \hspace{0.6cm}
\subfigure{\label{fig:Ficus_HoloLens_intern_pose}  
     \includegraphics[width=0.145\linewidth]{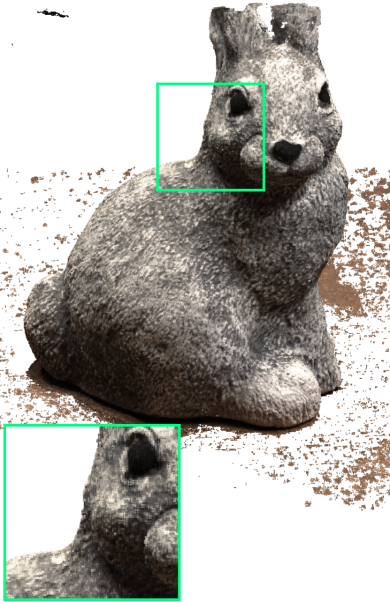}}\hspace{0.6cm}
\subfigure{\label{fig:Ficus_HoloLens_COLMAP_pose}  
     \includegraphics[width=0.145\linewidth]{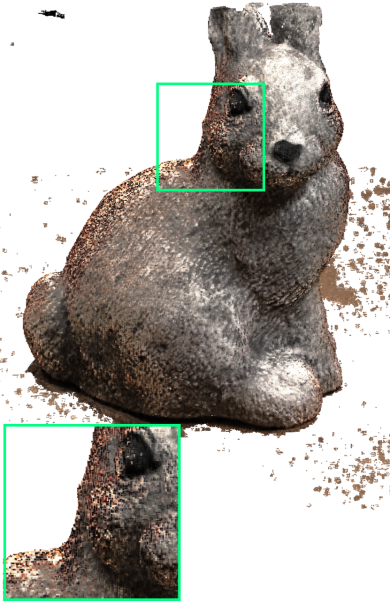}}\hspace{0.6cm}
\subfigure{\label{fig:Ficus_HoloLens_COLMAP_MVS}
	\includegraphics[width=0.145\linewidth]{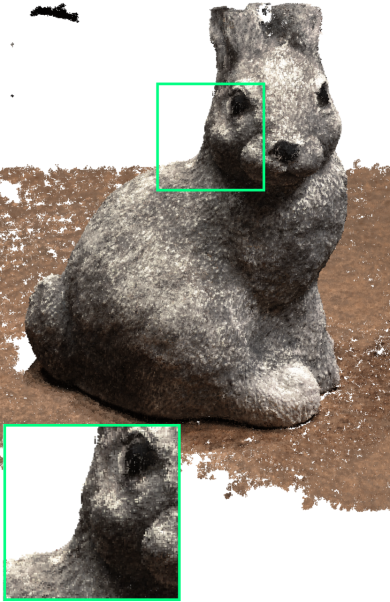}}\hspace{0.6cm}
\subfigure{\label{fig:Ficus_HoloLens_COLMAP_MVS}
	\includegraphics[width=0.145\linewidth]{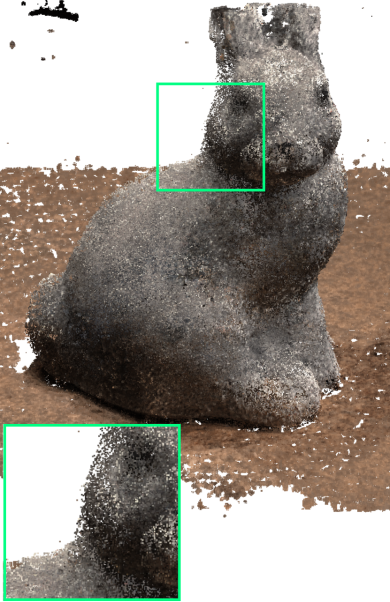}}\\
	\rotatebox{90}{$\,\,\,\,\,\,\,\,\,\,\,\,\,\,\,\,$scan63}
	\hspace{0.2cm}
\subfigure{\label{fig:Ficus_HoloLens_intern}
	\includegraphics[width=0.145\linewidth]{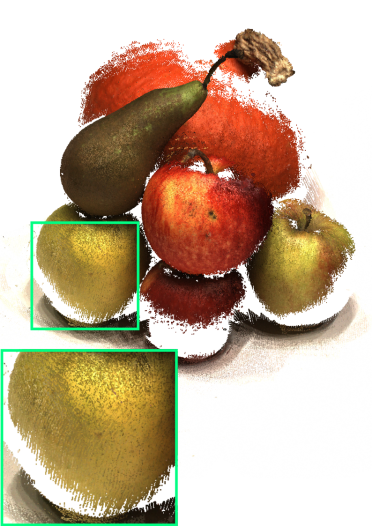}}\hspace{0.57cm}
\subfigure{\label{fig:Ficus_HoloLens_intern_pose}  
     \includegraphics[width=0.145\linewidth]{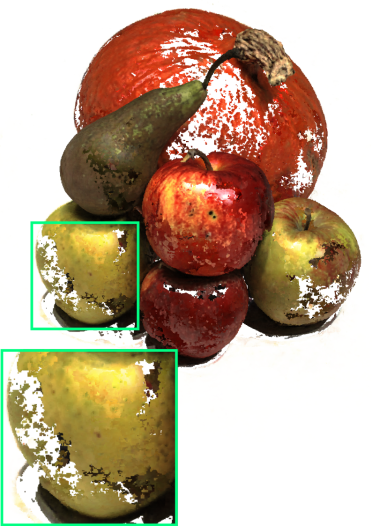}} \hspace{0.57cm}
\subfigure{\label{fig:Ficus_HoloLens_COLMAP_pose}  
     \includegraphics[width=0.145\linewidth]{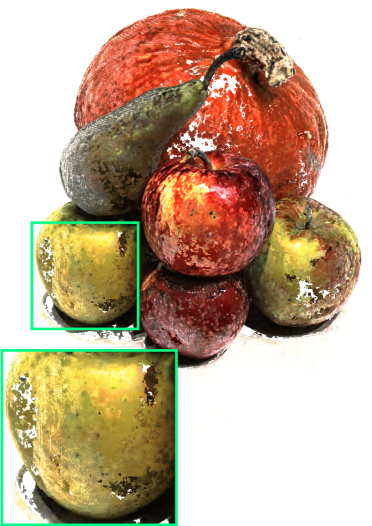}}\hspace{0.57cm}
\subfigure{\label{fig:Ficus_HoloLens_COLMAP_MVS}
	\includegraphics[width=0.145\linewidth]{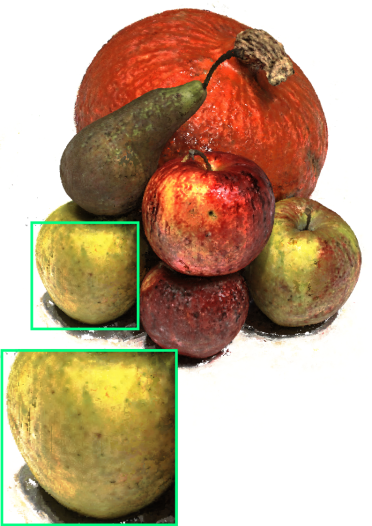}}\hspace{0.57cm}
\subfigure{\label{fig:Ficus_HoloLens_COLMAP_MVS}
	\includegraphics[width=0.145\linewidth]{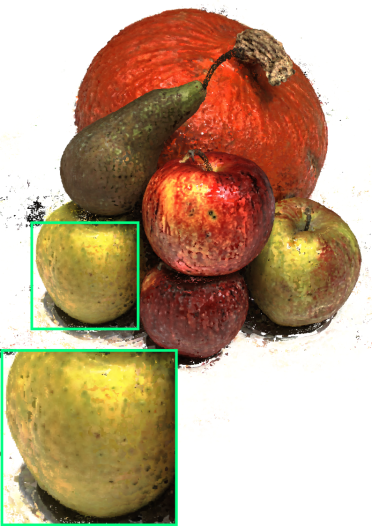}}\\
\rotatebox{90}{$\,\,\,\,\,\,\,\,\,\,\,\,\,\,\,\,$scan114}
\hspace{0.2cm}
\subfigure{\label{fig:Ficus_HoloLens_intern}
	\includegraphics[width=0.145\linewidth]{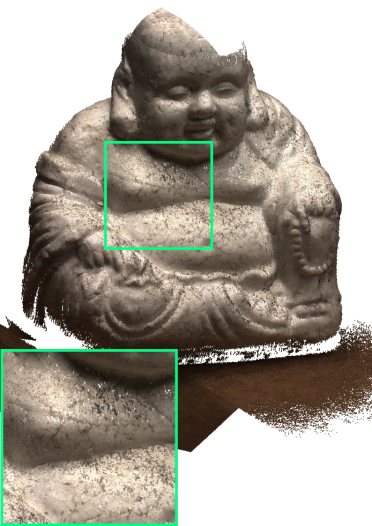}} \hspace{0.45cm}
\subfigure{\label{fig:Ficus_HoloLens_intern_pose}  
     \includegraphics[width=0.145\linewidth]{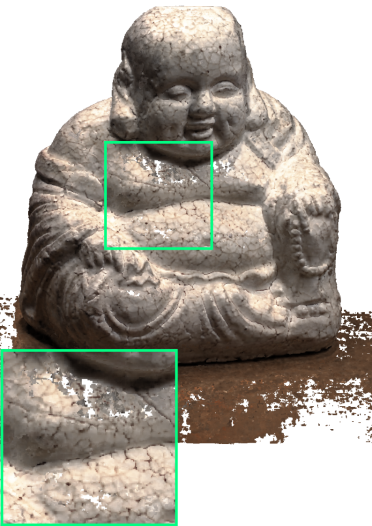}} \hspace{0.45cm}
\subfigure{\label{fig:Ficus_HoloLens_COLMAP_pose}  
     \includegraphics[width=0.145\linewidth]{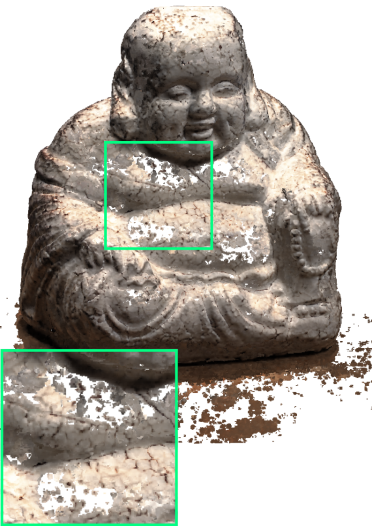}}\hspace{0.45cm}
\subfigure{\label{fig:Ficus_HoloLens_COLMAP_MVS}
	\includegraphics[width=0.145\linewidth]{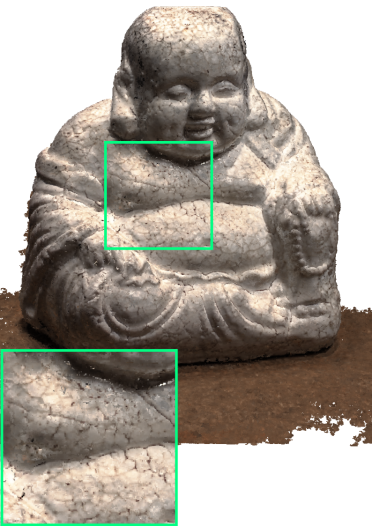}}\hspace{0.45cm}
\subfigure{\label{fig:Ficus_HoloLens_COLMAP_MVS}
	\includegraphics[width=0.145\linewidth]{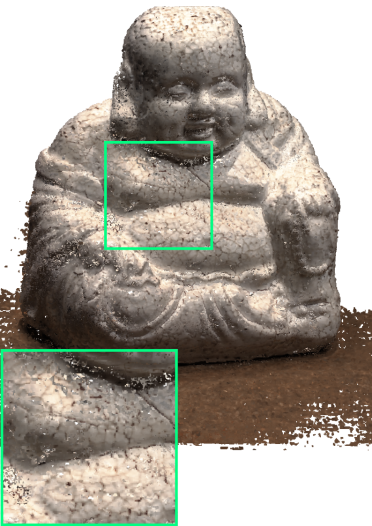}}
	\vspace{-3mm}
	\caption[Qualitative comparison on the real DTU benchmark dataset. Comparison between the reference point clouds and the geometric reconstructions from 3D density field using a global density threshold $\delta_{\text{t=50}}$, density gradients from Sobel filter $\Delta_{\delta,\text{Sobel}}$, Canny filter $\Delta_{\delta,\text{Canny}}$ and Laplacian of Gaussian $\Delta^2_{\delta,\text{LOG}}$.]
	{Qualitative comparison on the real DTU benchmark dataset. Comparison between the reference point clouds and the geometric reconstructions from 3D density field using a global density threshold $\delta_{\text{t=50}}$, density gradients from Sobel filter $\Delta_{\delta,\text{Sobel}}$, Canny filter $\Delta_{\delta,\text{Canny}}$ and Laplacian of Gaussian $\Delta^2_{\delta,\text{LOG}}$.}
\label{fig:pointclouds_dtu}
\end{figure*}

\begin{table*}
\caption{\textbf{Completeness}. Geometric completeness with the number of points $\uparrow$ in million of the reference with the geometric reconstructions from voxelized 3D density field with global density thresholds $\delta_{\text{t}}$, Sobel filter $\Delta_{\delta,\text{Sobel}}$, Canny filter $\Delta_{\delta,\text{Canny}}$ and Laplacian of Gaussian $\Delta^2_{\delta,\text{LOG}}$ within distance thresholds of a maximum distance of about 0.5\,mm, 1.0\,mm and 1.5\,mm from reference. The percentage $\uparrow$ is shown in brackets. Best results bold in \textcolor{green}{green}, second best results bold in \textcolor{blue}{blue}.} 
\label{tab:completeness}
\begin{tabularx}{\textwidth}{@{} l *{7}{r} @{}}
\toprule
 						   &scan24&scan37&scan40&scan55&scan63 & scan114 &  mean in $\%$ \\ \midrule
		 & &&&& &&\\
1.5\,mm	     		 & &&&& &&\\
$\delta_{\text{t=25}}$ 				  	&3.08 (\textcolor{green}{\textbf{96.99}}\,$\%$)&2.27 (\textcolor{green}{\textbf{97.58}}\,$\%$) &2.95 (\textcolor{blue}{\textbf{98.36}}\,$\%$)&2.96 (\textcolor{green}{\textbf{90.80}}\,$\%$)&1.10 (97.51\,$\%$) &3.02 (\textcolor{green}{\textbf{97.66}}\,$\%$) &  \textcolor{green}{\textbf{96.48}} \\	
$\delta_{\text{t=50}}$ 				   	&2.98 (93.79\,$\%$)	&2.10 (90.48\,$\%$) &2.86 (95.31\,$\%$)&2.35 (72.22\,$\%$)&0.87 (77.42\,$\%$) &2.79 (90.27\,$\%$) &  86.58 \\	
$\delta_{\text{t=100}}$ 				&2.79 (87.90\,$\%$)	&1.66 (71.54\,$\%$) &2.66 (88.51\,$\%$)&1.94 (59.53\,$\%$)&0.47 (41.71\,$\%$) &2.39 (77.17\,$\%$)& 71.07   \\		
$\Delta_{\delta,\text{Sobel}}$	&3.03 (\textcolor{blue}{\textbf{95.24}}\,$\%$)&2.03 (87.30\,$\%$) &2.88 (96.01\,$\%$)&2.09 (63.98\,$\%$)&1.02 (90.83\,$\%$) &2.40 (77.48\,$\%$) & 85.14 \\	
$\Delta_{\delta,\text{Canny}}$	&3.00 (94.43\,$\%$)	&2.26 (\textcolor{blue}{\textbf{97.28}}\,$\%$) &2.97 (\textcolor{green}{\textbf{98.93}}\,$\%$)&2.90 (\textcolor{blue}{\textbf{88.91}}\,$\%$)&1.12 (\textcolor{green}{\textbf{99.49}}\,$\%$) &3.02 (\textcolor{blue}{\textbf{97.54}}\,$\%$) &  \textcolor{blue}{\textbf{96.10}}\\	
$\Delta^2_{\delta,\text{LOG}}$	&2.73 (85.76\,$\%$)	&2.21 (95.08\,$\%$) &2.79 (93.11\,$\%$)&2.52 (77.15\,$\%$)&1.11 (\textcolor{blue}{\textbf{98.30}}\,$\%$) &2.95 (95.33\,$\%$)&  90.79 \\

	 & &&&& &&\\
1.0\,mm	     		 & &&&&  &&\\
$\delta_{\text{t=25}}$ 				  	&2.98 (\textcolor{green}{\textbf{93.60}}\,$\%$) &2.21 (\textcolor{green}{\textbf{94.89}}\,$\%$)&2.87 (\textcolor{green}{\textbf{95.80}}\,$\%$) &2.82 (\textcolor{green}{\textbf{86.43}}\,$\%$)&1.06 (\textcolor{blue}{\textbf{93.80}}\,$\%$) &2.95 (\textcolor{green}{\textbf{95.55}}\,$\%$)& \textcolor{green}{\textbf{93.35}} \\	
$\delta_{\text{t=50}}$ 				   	&2.82 (88.73\,$\%$) &1.92 (82.80\,$\%$)&2.71 (90.50\,$\%$) &2.17 (66.65\,$\%$)&0.75 (66.83\,$\%$) &2.64 (85.25\,$\%$)& 80.13 \\	
$\delta_{\text{t=100}}$ 				&2.52 (79.38\,$\%$) &1.41 (60.58\,$\%$)&2.43 (80.96\,$\%$) &1.86 (57.08\,$\%$)&0.36 (32.32\,$\%$) &2.11 (68.22\,$\%$)&  63.09 \\		
$\Delta_{\delta,\text{Sobel}}$	&2.94 (\textcolor{blue}{\textbf{92.43}}\,$\%$) &1.86 (80.17\,$\%$)&2.82 (\textcolor{blue}{\textbf{93.92}}\,$\%$) &2.02 (61.87\,$\%$)&0.96 (84.94\,$\%$) &2.23 (72.00\,$\%$)&  80.88 \\	
$\Delta_{\delta,\text{Canny}}$	&2.46 (77.22\,$\%$) &2.08 (\textcolor{blue}{\textbf{89.46}}\,$\%$)&2.73 (91.05\,$\%$) &2.27 (\textcolor{blue}{\textbf{69.57}}\,$\%$)&1.10 (\textcolor{green}{\textbf{98.06}}\,$\%$) &2.87 (\textcolor{blue}{\textbf{92.73}}\,$\%$) & \textcolor{blue}{\textbf{86.35}} \\	
$\Delta^2_{\delta,\text{LOG}}$	&2.13 (66.94\,$\%$) &1.93 (83.15\,$\%$)&2.17 (72.36\,$\%$) &1.98 (60.66\,$\%$)&1.03 (91.48\,$\%$) &2.57 (83.18\,$\%$)& 76.30  \\

	 & &&&&& &\\
0.5\,mm	     		 & &&& &&&\\
$\delta_{\text{t=25}}$				  	&2.63 (\textcolor{blue}{\textbf{82.86}}\,$\%$)&1.81 (\textcolor{green}{\textbf{77.70}}\,$\%$) &2.28 (\textcolor{blue}{\textbf{75.86}}\,$\%$)&2.20 (\textcolor{green}{\textbf{67.46}}\,$\%$)&0.92 (\textcolor{green}{\textbf{81.48}}\,$\%$) &2.73 (\textcolor{green}{\textbf{88.21}}\,$\%$)  & \textcolor{green}{\textbf{78.93}}  \\	
$\delta_{\text{t=50}}$ 				   	&2.31 (72.66\,$\%$)&1.39 (59.63\,$\%$) &2.00 (66.74\,$\%$)&1.60 (49.03\,$\%$)&0.55 (48.97\,$\%$) &2.22 (\textcolor{blue}{\textbf{71.66}}\,$\%$) & 61.45 \\	
$\delta_{\text{t=100}}$ 				&1.70 (53.37\,$\%$)&0.88 (37.83\,$\%$) &1.59 (53.03\,$\%$)&1.21 (37.15\,$\%$)&0.22 (19.48\,$\%$) &1.49 (48.03\,$\%$)&  41.48  \\		
$\Delta_{\delta,\text{Sobel}}$	&2.69 (\textcolor{green}{\textbf{84.67}}\,$\%$)&1.40  (\textcolor{blue}{\textbf{60.12}}\,$\%$) &2.43 (\textcolor{green}{\textbf{80.86}}\,$\%$)&1.77 (\textcolor{blue}{\textbf{54.22}}\,$\%$)&0.81 (72.39\,$\%$) &1.89 (61.00\,$\%$)&  \textcolor{blue}{\textbf{68.88}}\\	
$\Delta_{\delta,\text{Canny}}$	&1.14 (35.85\,$\%$)&1.00 (42.86\,$\%$) &1.25 (41.69\,$\%$)&0.80 (24.67\,$\%$)&0.87 (\textcolor{blue}{\textbf{77.51}}\,$\%$) &1.96 (63.55\,$\%$) & 47.69 \\	
$\Delta^2_{\delta,\text{LOG}}$	&1.20 (37.66\,$\%$)&1.13 (48.73\,$\%$) &0.81 (26.85\,$\%$)&1.16 (35.59\,$\%$)&0.75 (66.99\,$\%$) &1.58 (51.19\,$\%$) &   44.50\\	
	 
\bottomrule
\end{tabularx}
\end{table*}

\section{EXPERIMENTS}\label{sec:experiments}
In this section, we conduct experiments on a challenging benchmark dataset with different types of real objects, which feature different sizes, structures, materials, textures and colors.

\subsection{Dataset}
For the evaluation of our framework, we use the DTU benchmark dataset \citep{dtu}. The dataset consists of scenes featuring real objects, including images, corresponding camera poses, and reference point clouds obtained from a structured-light scanner (SLS). We specifically focus on six scenes within the dataset, the same as \citep{NeuS,unisurf,NeuralWarp,neuralangelo}, each containing either 49 or 64 RGB images.

\subsection{Implementation} For all investigations, Instant NGP \citep{mueller_et_al_2022} was taken into account as NeRF, since it enables real time training and rendering. Regarding the network architecture, the basic NeRF architecture with ReLu activations and hash encoding is selected, while the training incorporates 50\,000 training steps on an NVIDIA RTX3090 GPU.

\subsection{Experiments}

We evaluate our framework with first derivative Sobel filter and Canny filter as well as second derivative Laplacian of Gaussian filter against different global density thresholds. Thereby qualitative as well as quantitative results based on completeness and correctness as described in the evaluation Section \ref{sec:evaluation} are considered.
The global density thresholds $\delta_{\text{t}}$ are set to 25, 50, and 100 \citep{NeuS}. The Sobel filter is used as described in Section \ref{sec:First_derivative} and the Canny filter is applied with a standard derivation of 0.1 and relative thresholds of 0.0005 and 4 times 0.0005. For the Laplacian of Gaussian, a filter mask of 7$\times$7$\times$7 and standard derivation of 7 is utilized.


\section{RESULTS}
In the following sections, we show qualitative (Section \ref{sec:results_qualitative}) and quantitative (Section \ref{sec:results_quanitative}) results of the geometric reconstructions on the used benchmark dataset by addressing completeness and correctness. 

\subsection{Qualitative results}\label{sec:results_qualitative}
As the following Figures \ref{fig:pointclouds_dtu} and \ref{fig:pointclouds_dtu_p2p} show, the density gradient-based approach with 3D edge detection filters yields promising results. Thus, the optimal global density threshold varies from scene to scene and requires adaptive adjustment.
By addressing the density gradient, consistently accurate and complete results are generated across all scenes.

\paragraph{Completeness}
The visual comparison of the colored geometric reconstructions in Figure \ref{fig:pointclouds_dtu} highlights the reconstruction quality and object completeness based on density gradients. The reconstructions exemplified for a global density threshold $\delta_{\text{t=50}}$ exhibit different levels of gaps in the point clouds. In almost all scenes, gaps appear in the reconstructions along with areas of extremely high point density. Also the reconstructions resulting from the first derivative, the Sobel filter $\Delta_{\delta,\text{Sobel}}$, performs slightly different depending on the scene. For certain scenes like scan40 and scan55 the detected edges seem to be located too far above the reference surface. For the other scenes, however, substantial gaps exist. The Canny filter $\Delta_{\delta,\text{Canny}}$ provides the strongest visual results. Besides the colorful and smooth objects like scene scan63, also complex collections like in scene scan37 can be reconstructed almost completely. In general, the Canny filter effectively eliminates gaps and delivers a uniform point density. In addition, the subsurface of the scenes containing a colored ground are well captured. The second derivation with the Laplacian of Gaussian $\Delta^2_{\delta,\text{LOG}}$ also reaches a complete reconstruction at first sight. However, especially at scene scan55 and partly scene scan40 the point cloud tends to be rather fuzzy and noisy.

\paragraph{Correctness}
Besides the visually strong results, the density gradient applications also provide geometrically promising results (Figure \ref{fig:pointclouds_dtu_p2p}).
Both the global density threshold as well as the gradient filters enable results with accuracies up to 2.5\,mm for most parts of the object surfaces. Nevertheless, especially in reconstructions from global density thresholds and Sobel filter $\Delta_{\delta,\text{Sobel}}$, some artifacts up to 10\,mm appear. In contrast, the Canny filter $\Delta_{\delta,\text{Canny}}$ provides mainly consistent high accuracies to about 1.5\,mm. The geometric accuracy of the Laplacian of Gaussian $\Delta^2_{\delta,\text{LOG}}$ depends highly on the scene and appears quite noisy.
Note that edged areas with a large deviation are mainly due to the missing parts of the points in the SLS reference, which are depicted in the images and therefore in the reconstruction from 3D density field.

\subsection{Quantitative results}\label{sec:results_quanitative}

\paragraph{Completeness}

The reported completeness reached by the different NeRF reconstructions is shown in Table \ref{tab:completeness} and specified by absolute number of points as well as percentage.
Altogether, the completeness values for the distance thresholds up to 1 and 1.5\,mm exceed 60\,$\%$ for all methods. As expected, an increase of the density threshold $\delta_{\text{t}}$ causes a decrease of the completeness, due to the fact that a higher number of points are removed. Accordingly, using a density threshold of 25 results in the highest completeness for this dataset, with a mean across scenes of approximately 96\,$\%$ for points below 1.5\,mm and 93\,$\%$ up to 1\,mm accuracy. The results from density gradients through the Canny filter also stands out strongly. On average, 96\,$\%$ completeness is achieved for 1.5\,mm and 86\,$\%$ for 1\,mm.
Taking a more detailed look, the methods perform variably for each scene. 
Canny performs particularly well on complex, specular and smooth objects as in scene scan37, scan63 or scan114. However, at fine detail levels and rough objects like in scene scan24 and scan55, the completeness quickly weakens for highly accurate reconstructions up to 0.5\,mm. Nonetheless, the completeness decreases significantly using global density thresholds starting from $\delta_{\text{t=50}}$ and can not compete with the Canny filter.

\begin{figure}[H]
\centering
\subfigure[]{\label{fig:full}\includegraphics[width=0.15\textwidth]{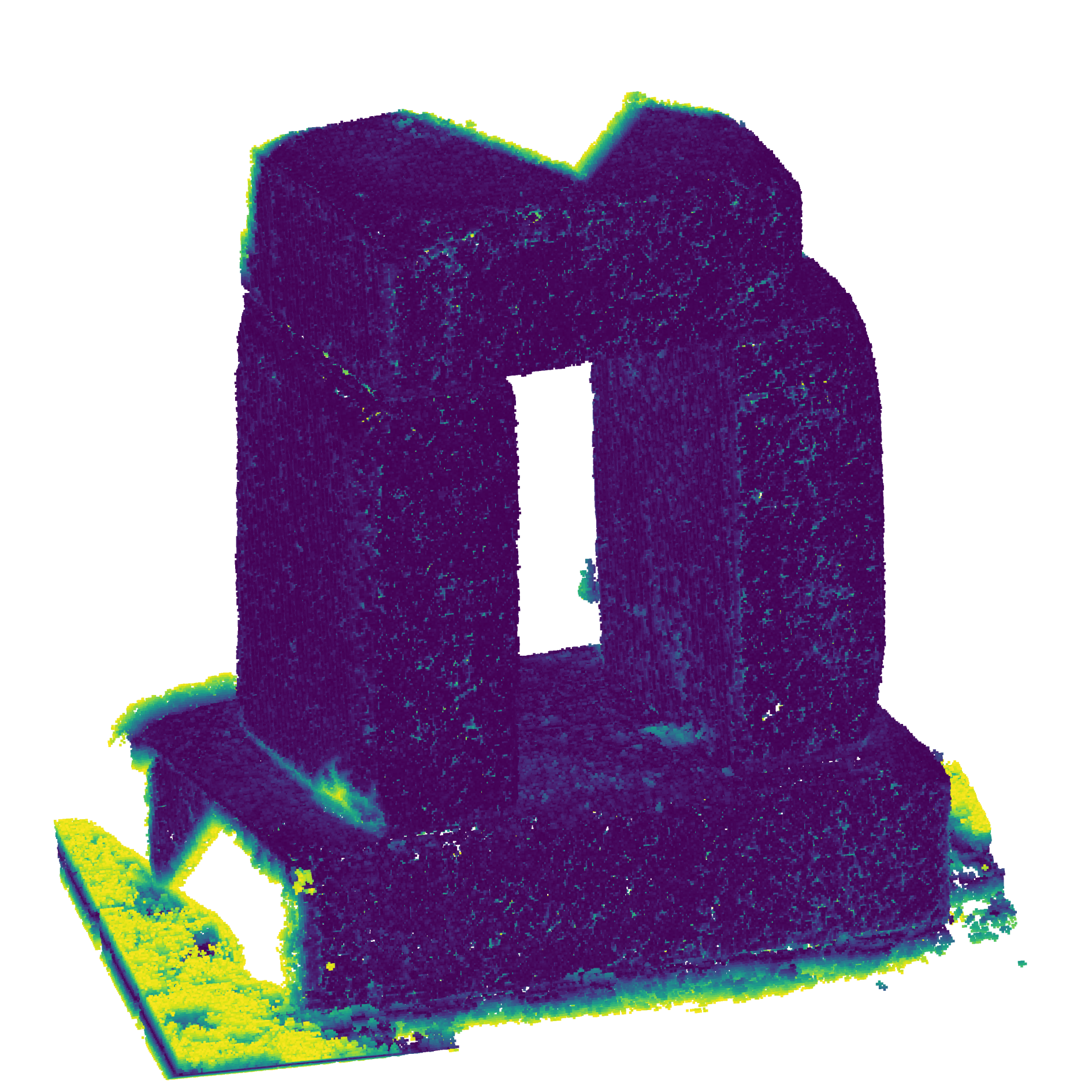}}
\hspace{0.5cm}\subfigure[]{\label{fig:15}\includegraphics[width=0.15\textwidth]{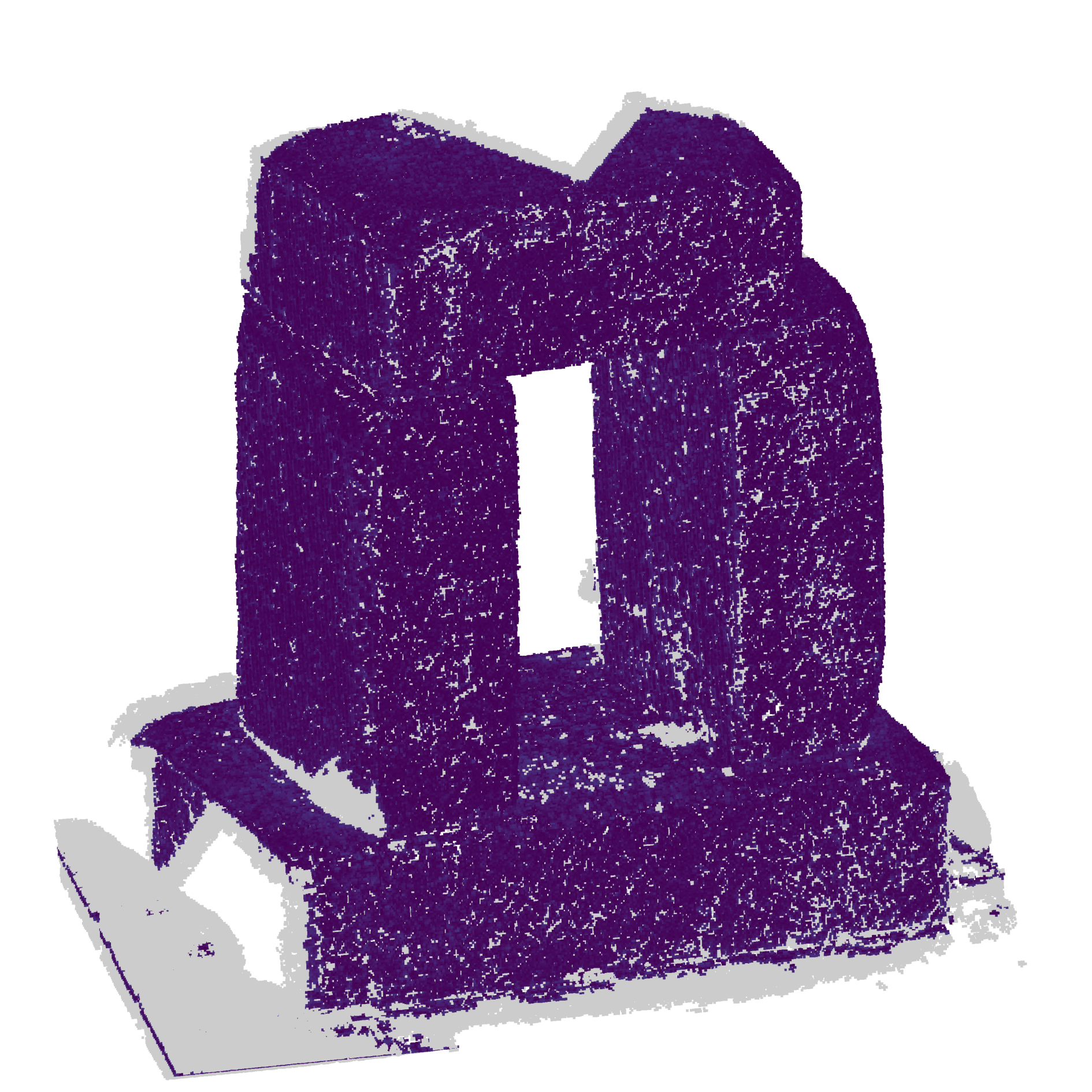}}
\subfigure[]{\label{fig:10}\includegraphics[width=0.15\textwidth]{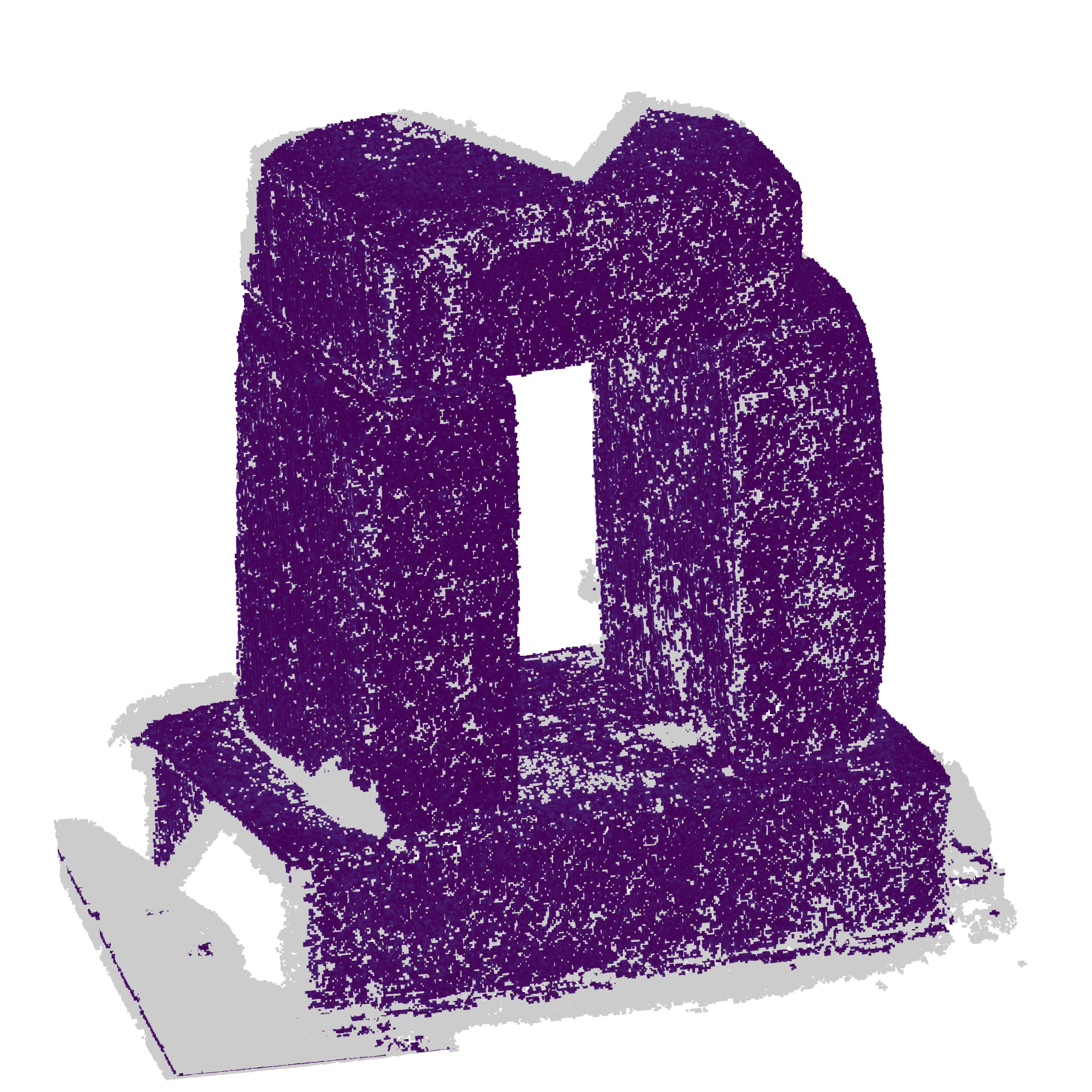}}
\hspace{0.5cm}\subfigure[]{\label{fig:05}\includegraphics[width=0.15\textwidth]{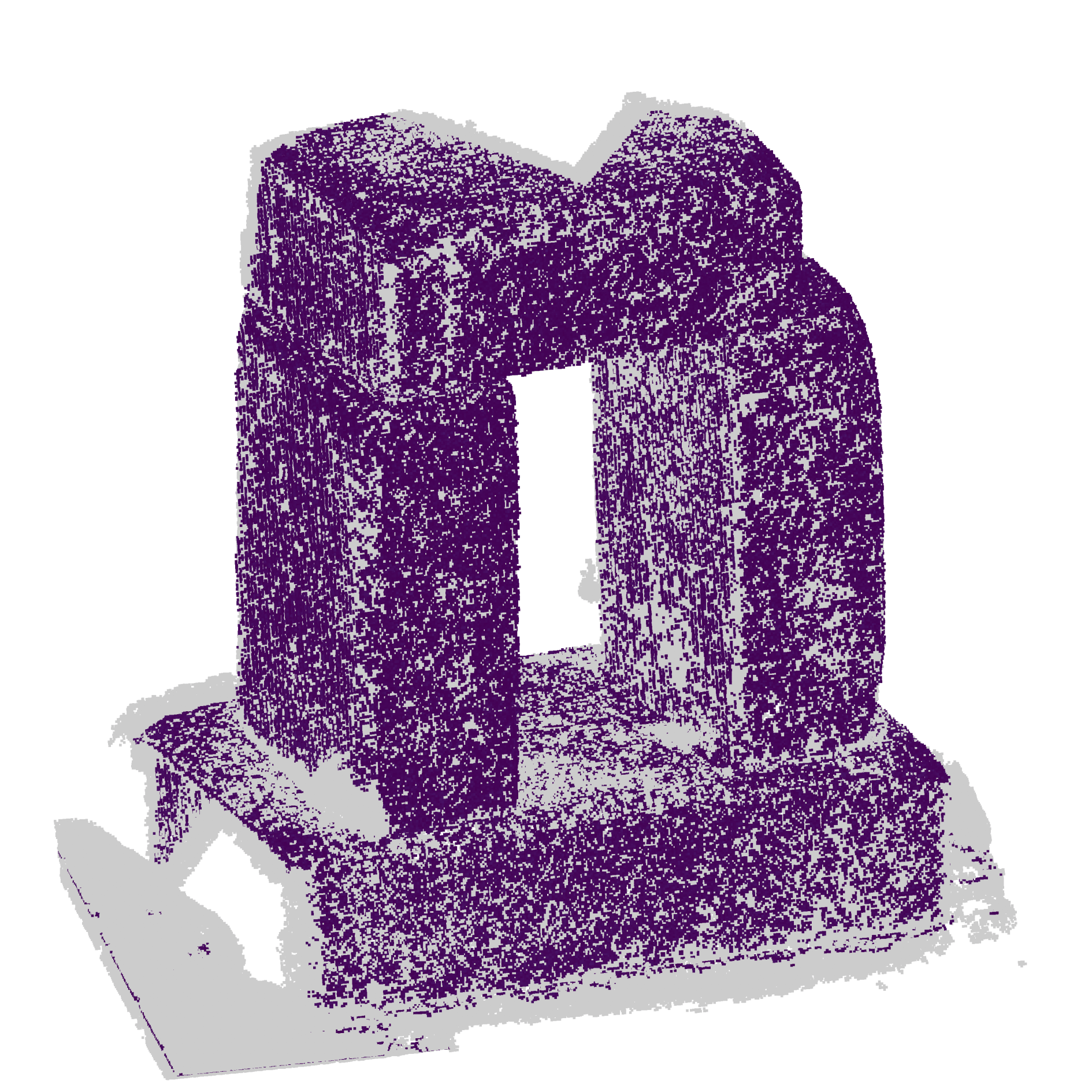}}\vspace{-5mm}
\caption[Extraction of surface points (colored) of a reconstruction using Canny filter up to a geometric accuracy (data-to-reference) \protect{\subref{fig:full}} 10.0\,mm, \protect{\subref{fig:15}} 1.5\,mm, \protect{\subref{fig:10}} 1.0\,mm, and \protect{\subref{fig:05}} 0.5\,mm. Points beyond are grey. The surface points show a high spatial density combined with a high geometric accuracy.]{Extraction of surface points (colored) of a reconstruction using Canny filter below a accuracy of \protect{\subref{fig:full}} 10.0\,mm, \protect{\subref{fig:15}} 1.5\,mm, \protect{\subref{fig:10}} 1.0\,mm, and \protect{\subref{fig:05}} 0.5\,mm. Points beyond are grey. The surface points show a high spatial density combined with a high geometric accuracy.}
\label{fig:c2c_values}
\end{figure}

\paragraph{Correctness}
Since NeRFs estimate values in the entire 3D space and consequently inside the objects, there may be artifactual points within the objects, which affect the quantitative accuracy results in terms of correctness. 
Respectively, the results (Table \ref{tab:correctness}) are mostly in the same, rather coarse, range of accuracies up to 6\,mm, from the NeRF reconstructions to the reference (data-to-reference). While using a global threshold performs differently depending on the scene, using density gradients remains consistently stable across all scenes. Nevertheless, the points within the object undermine the interpretability of the results. 
To emphasize the accuracy potential of density gradients, considering the reconstruction surface points, Figure \ref{fig:c2c_values} shows the surface points below 0.5, 1.0 and 1.5\,mm for a result based on the Canny filter. It illustrates that the Canny filter approach generates a densely sampled scene whose surface points exhibit high geometric accuracies.
When viewing the accuracy from the reference to the 3D reconstructions (reference-to-data), the density gradients stand out positively with an achieved correctness compared to the global density thresholds as well.

\begin{figure*}[h!]
	\centering
\raisebox{\dimexpr 0cm-\height}{reference}
\hspace{2cm}
\raisebox{\dimexpr 0cm-\height}{$\delta_{\text{t=50}}$}
\hspace{2.5cm}
\raisebox{\dimexpr 0cm-\height}{$\Delta_{\delta,\text{Sobel}}$}
\hspace{2.5cm}
\raisebox{\dimexpr 0cm-\height}{$\Delta_{\delta,\text{Canny}}$}
\hspace{2cm}
\raisebox{\dimexpr 0cm-\height}{$\Delta^2_{\delta,\text{LOG}}$}
\hspace{3cm}\\
\rotatebox{90}{scan24}
\vspace{0.2mm}
\subfigure{\label{fig:Ficus_HoloLens_intern}
	\includegraphics[width=0.38\columnwidth]{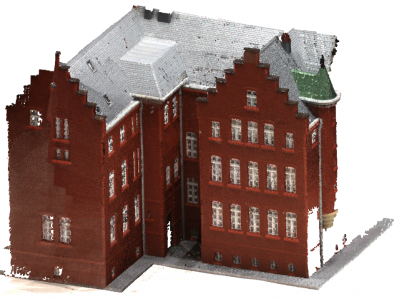}} 
\subfigure{\label{fig:Ficus_HoloLens_intern_pose}  
     \includegraphics[width=0.38\columnwidth]{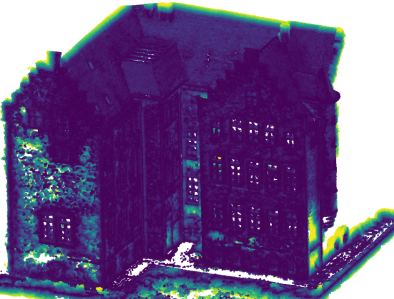}} 
\subfigure{\label{fig:Ficus_HoloLens_COLMAP_pose}  
     \includegraphics[width=0.38\columnwidth]{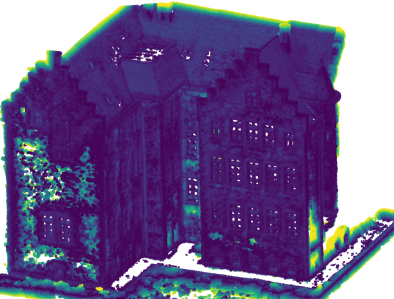}}
\subfigure{\label{fig:Ficus_HoloLens_COLMAP_MVS}
	\includegraphics[width=0.38\columnwidth]{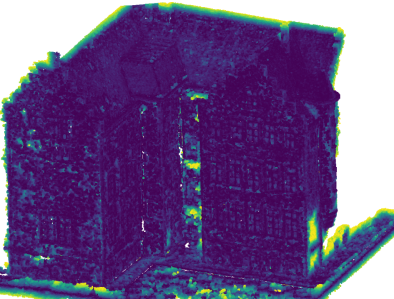}}
\subfigure{\label{fig:Ficus_HoloLens_COLMAP_MVS}
	\includegraphics[width=0.38\columnwidth]{result_images_new/scan24_canny_p2p.png}}\\
\rotatebox{90}{scan37}
\vspace{0.2mm}
\subfigure{\label{fig:Ficus_HoloLens_intern}
	\includegraphics[width=0.38\columnwidth]{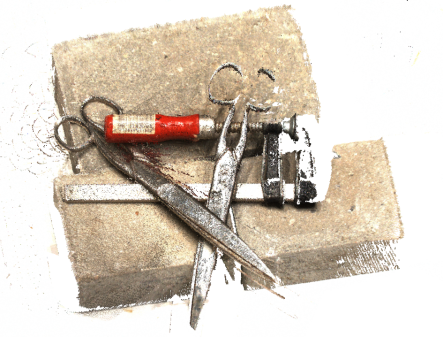}} 
\subfigure{\label{fig:Ficus_HoloLens_intern_pose}  
     \includegraphics[width=0.38\columnwidth]{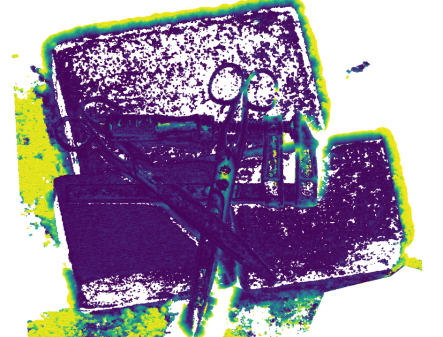}} 
\subfigure{\label{fig:Ficus_HoloLens_COLMAP_pose}  
     \includegraphics[width=0.38\columnwidth]{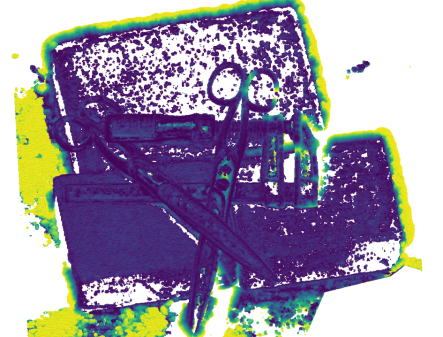}}
\subfigure{\label{fig:Ficus_HoloLens_COLMAP_MVS}
	\includegraphics[width=0.38\columnwidth]{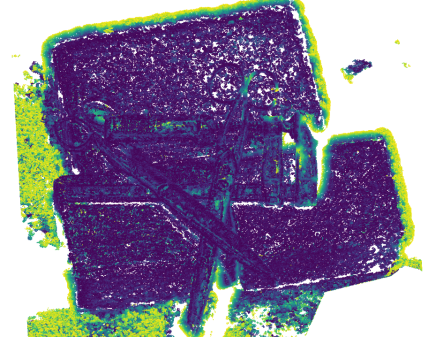}}
\subfigure{\label{fig:Ficus_HoloLens_COLMAP_MVS}
	\includegraphics[width=0.38\columnwidth]{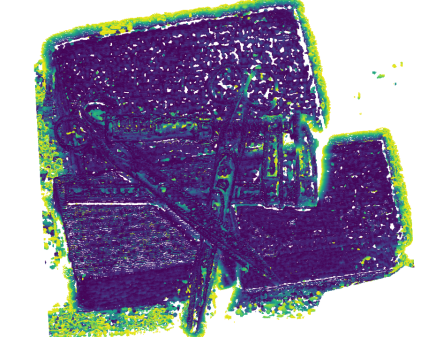}}\\
\rotatebox{90}{scan40}
\subfigure{\label{fig:Ficus_HoloLens_intern}
	\includegraphics[width=0.38\columnwidth]{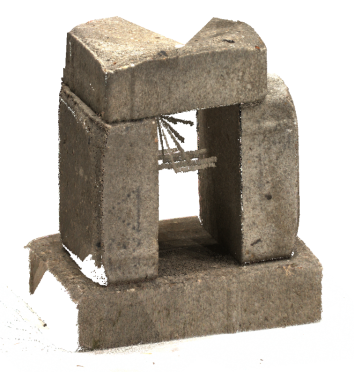}} 
\subfigure{\label{fig:Ficus_HoloLens_intern_pose}  
     \includegraphics[width=0.38\columnwidth]{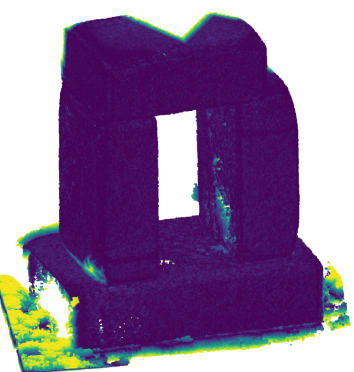}} 
\subfigure{\label{fig:Ficus_HoloLens_COLMAP_pose}  
     \includegraphics[width=0.38\columnwidth]{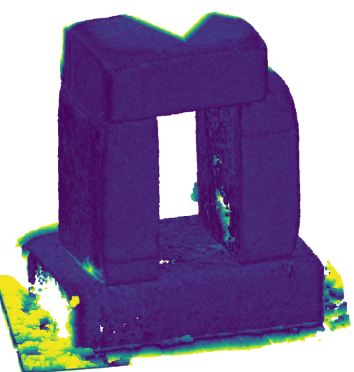}}
\subfigure{\label{fig:Ficus_HoloLens_COLMAP_MVS}
	\includegraphics[width=0.38\columnwidth]{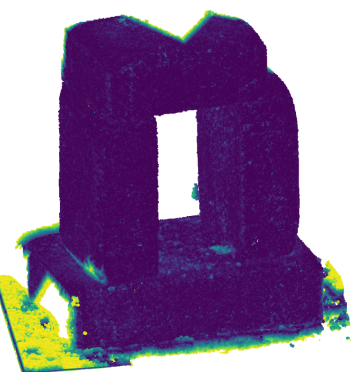}}
	\subfigure{\label{fig:Ficus_HoloLens_COLMAP_MVS}
	\includegraphics[width=0.38\columnwidth]{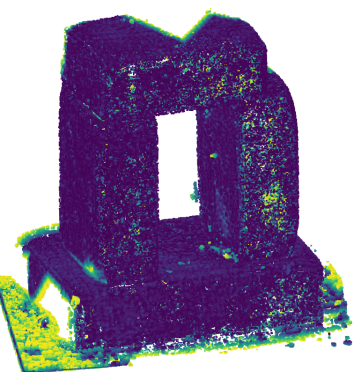}}\\
\rotatebox{90}{scan55}
\hspace{0.3cm}
\subfigure{\label{fig:Ficus_HoloLens_intern}
	\includegraphics[width=0.37\columnwidth]{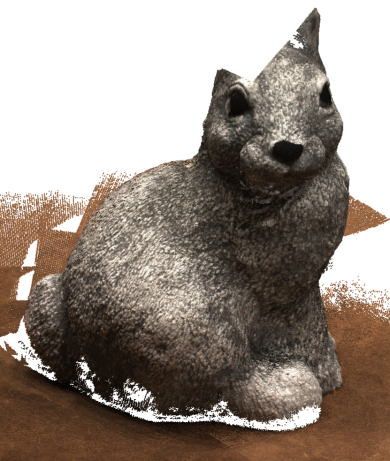}} 
\subfigure{\label{fig:Ficus_HoloLens_intern_pose}  
     \includegraphics[width=0.37\columnwidth]{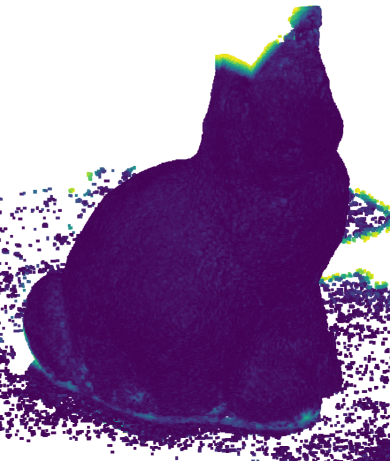}} 
\subfigure{\label{fig:Ficus_HoloLens_COLMAP_pose}  
     \includegraphics[width=0.37\columnwidth]{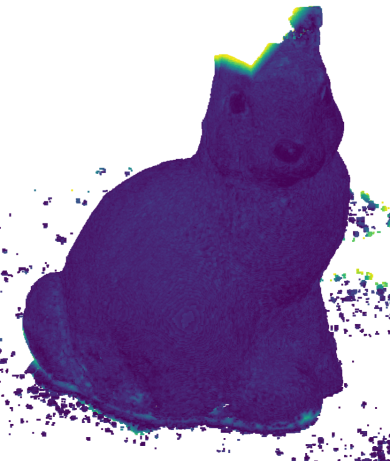}}
\subfigure{\label{fig:Ficus_HoloLens_COLMAP_MVS}
	\includegraphics[width=0.37\columnwidth]{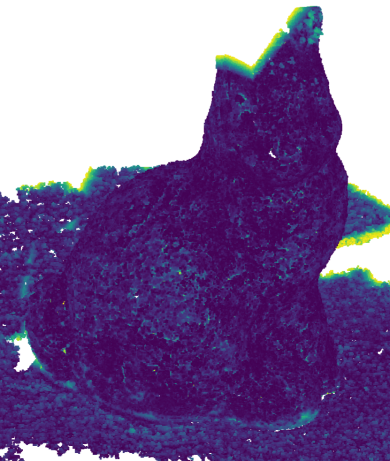}}
\subfigure{\label{fig:Ficus_HoloLens_COLMAP_MVS}
	\includegraphics[width=0.37\columnwidth]{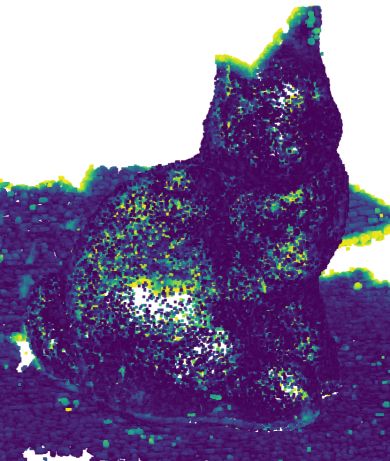}}\\
\rotatebox{90}{scan63}
\hspace{0.3cm}
\subfigure{\label{fig:Ficus_HoloLens_intern}
	\includegraphics[width=0.37\columnwidth]{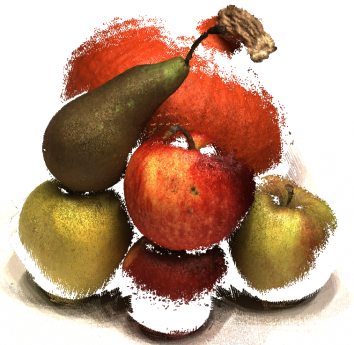}} 
\subfigure{\label{fig:Ficus_HoloLens_intern_pose}  
     \includegraphics[width=0.37\columnwidth]{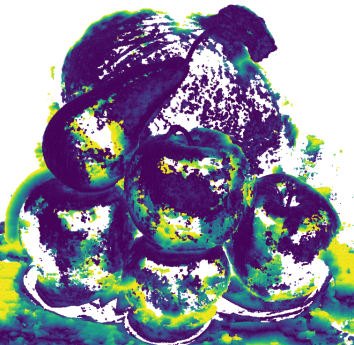}} 
\subfigure{\label{fig:Ficus_HoloLens_COLMAP_pose}  
     \includegraphics[width=0.37\columnwidth]{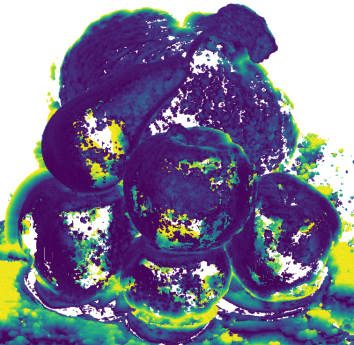}}
\subfigure{\label{fig:Ficus_HoloLens_COLMAP_MVS}
	\includegraphics[width=0.37\columnwidth]{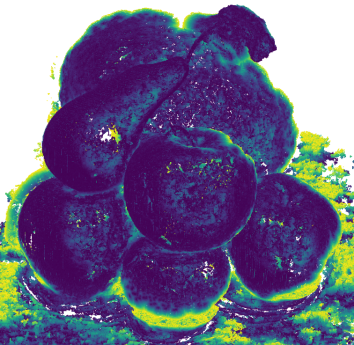}}
\subfigure{\label{fig:Ficus_HoloLens_COLMAP_MVS}
	\includegraphics[width=0.37\columnwidth]{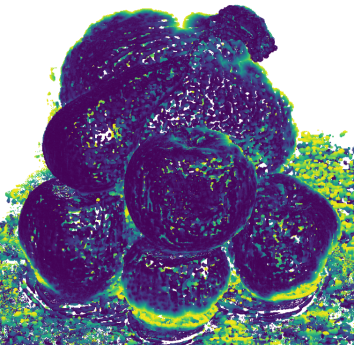}}\\
\rotatebox{90}{scan114}
\hspace{0.3cm}
\subfigure{\label{fig:Ficus_HoloLens_intern}
	\includegraphics[width=0.37\columnwidth]{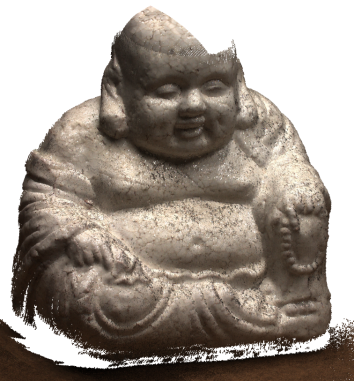}} 
\subfigure{\label{fig:Ficus_HoloLens_intern_pose}  
     \includegraphics[width=0.37\columnwidth]{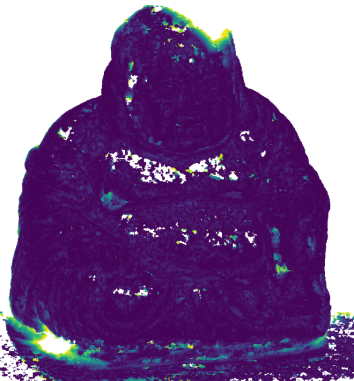}} 
\subfigure{\label{fig:Ficus_HoloLens_COLMAP_pose}  
     \includegraphics[width=0.37\columnwidth]{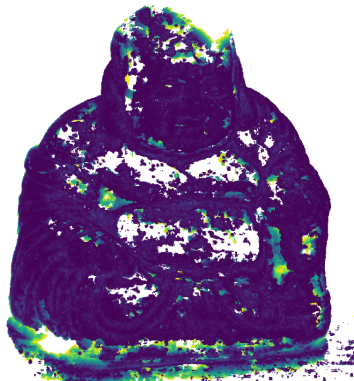}}
\subfigure{\label{fig:Ficus_HoloLens_COLMAP_MVS}
	\includegraphics[width=0.37\columnwidth]{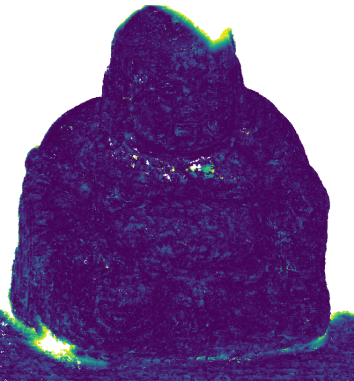}}
\subfigure{\label{fig:Ficus_HoloLens_COLMAP_MVS}
	\includegraphics[width=0.37\columnwidth]{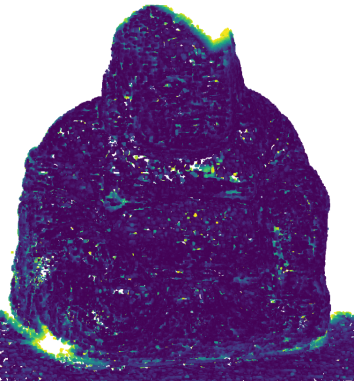}}\\
\subfigure{\label{fig:Ficus_HoloLens_COLMAP_MVS}
	\includegraphics[width=0.75\columnwidth]{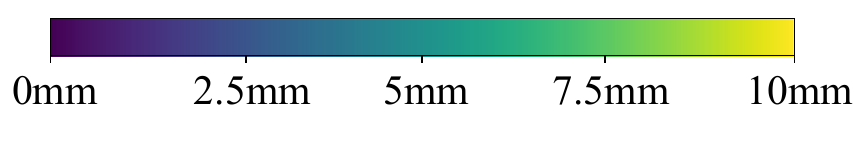}}
	\vspace{-5mm}
	\caption[Qualitative comparison on the real DTU benchmark dataset with Chamfer cloud-to-cloud distances. Comparison between the reference point clouds and the geometric reconstructions from 3D density field using a global density threshold $\delta_{\text{t=50}}$, density gradients from Sobel filter $\Delta_{\delta,\text{Sobel}}$, Canny filter $\Delta_{\delta,\text{Canny}}$ and Laplacian of Gaussian $\Delta^2_{\delta,\text{LOG}}$.]
	{Qualitative comparison on the real DTU benchmark dataset with Chamfer cloud-to-cloud distances. Comparison between the reference point clouds and the geometric reconstructions from 3D density field using a global density threshold $\delta_{\text{t=50}}$, density gradients from Sobel filter $\Delta_{\delta,\text{Sobel}}$, Canny filter $\Delta_{\delta,\text{Canny}}$ and Laplacian of Gaussian $\Delta^2_{\delta,\text{LOG}}$.}
\label{fig:pointclouds_dtu_p2p}
\vspace{1cm}
\end{figure*}
\begin{table*}
\caption{\textbf{Correctness}. Geometric accuracy with Chamfer cloud-to-cloud distance $\downarrow$ in mm of the geometric reconstructions from voxelized 3D density field with global density thresholds $\delta_{\text{t}}$, Sobel filter $\Delta_{\delta,\text{Sobel}}$, Canny filter $\Delta_{\delta,\text{Canny}}$ and Laplacian of Gaussian $\Delta^2_{\delta,\text{LOG}}$. From data-to-reference as well as reference-to-data. For comparable results, we use the same random permuted 2\,500\,000 points, since the resulting reconstructions include different numbers of points and consequently spatial point density. Best results bold in \textcolor{green}{green}, second best results bold in \textcolor{blue}{blue}.}
\label{tab:correctness}
\begin{tabularx}{\textwidth}{@{} l *{8}{>{\raggedleft\arraybackslash}X} @{}}
\toprule
 							   &scan24&scan37&scan40&scan55& scan63 & scan114 && \,\,\,\,\,mean\\
 					\midrule
data-to-reference (mm)   & &&&&& \\
$\delta_{\text{t=25}}$ 				   & 3.26 & 5.31 & 3.53 &3.02 & 5.93 &3.21 &	&\,\,\,\,\,4.04\\
$\delta_{\text{t=50}}$ 				   & \textcolor{blue}{\textbf{2.86}} & 5.63 & 3.44 &3.34& 6.08 &2.84 && \,\,\,\,\,4.03\\
$\delta_{\text{t=100}}$ 			   & 3.08 & 6.05 & \textcolor{blue}{\textbf{3.35}} 	&3.48& 6.25 &3.28	&& \,\,\,\,\,4.25\\
$\Delta_{\delta,\text{Sobel}}$  & \textcolor{green}{\textbf{2.79}} & 5.65 & \textcolor{green}{\textbf{3.31}}	&3.12& 6.20 &\textcolor{blue}{\textbf{2.72}} 	& &\,\,\,\,\,3.97\\
$\Delta_{\delta,\text{Canny}}$  &3.03	& \textcolor{blue}{\textbf{5.26}} & 3.62	&\textcolor{blue}{\textbf{2.76}}& \textcolor{blue}{\textbf{5.87}}	&2.85 && \textcolor{blue}{\textbf{\,\,\,\,\,3.89}}\\ 
$\Delta^2_{\delta,\text{LOG}}$  & 3.40 & \textcolor{green}{\textbf{3.98}} & 3.44	&\textcolor{green}{\textbf{2.06}}& \textcolor{green}{\textbf{5.65}}	&\textcolor{green}{\textbf{2.57}} 	&& \textcolor{green}{\textbf{\,\,\,\,\,3.52}}\\\midrule
reference-to-data (mm)  & &&&&& &\\ 
$\delta_{\text{t=25}}$ 				   & 0.82 & \textcolor{green}{\textbf{0.64}} & \textcolor{green}{\textbf{0.76}} 	&0.64&  0.82& 1.95	 &&\,\,\,\,\,0.94  \\
$\delta_{\text{t=50}}$ 				   & \textcolor{blue}{\textbf{0.79}} & 0.82 & 0.81 	&\textcolor{blue}{\textbf{0.60}}&  1.50& \textcolor{green}{\textbf{0.63}}	 && \,\,\,\,\,0.86 \\
$\delta_{\text{t=100}}$ 			   & 1.05 & 1.38 & 0.99 	&0.63&  3.97& 2.43 &&	\,\,\,\,\,1.74\\
$\Delta_{\delta,\text{Sobel}}$  & \textcolor{green}{\textbf{0.76}} & 0.96 & \textcolor{blue}{\textbf{0.77}}	&\textcolor{green}{\textbf{0.57}}& 1.02& 0.77  && \textcolor{blue}{\textbf{\,\,\,\,\,0.81}} \\
$\Delta_{\delta,\text{Canny}}$  & 0.91	& \textcolor{blue}{\textbf{0.67}} & 0.79 	&1.02&  \textcolor{green}{\textbf{0.70}}& \textcolor{blue}{\textbf{0.70}}  && \textcolor{green}{\textbf{\,\,\,\,\,0.80}}\\
$\Delta^2_{\delta,\text{LOG}}$  & 1.09 & 0.68 & 0.95 &1.27& \textcolor{blue}{\textbf{0.74}}& 0.88  &&\,\,\,\,\,0.93\\
\bottomrule
\end{tabularx}
\end{table*}

\newpage
\vspace{15cm}
\section{DISCUSSION}

In this paper, we investigate the density gradients for achieving high geometric completeness and correctness in 3D reconstructions based on density gradients from NeRFs density output. The application of gradient filters on the density field for 3D edge detection shows remarkable results, compared to the usage of global density thresholds. The latter often leads to gaps or artifacts in the reconstructions, depending on the chosen threshold. However, by extracting surfaces based on density gradients, we can overcome this issue.


The qualitative results of the density gradients, especially by the Canny filter, consistently stand out as positive over all scenes in terms of completeness, this aligns with both quantitative and qualitative results.
While global density thresholds yield good results, scene-dependent accuracy variations exist. For some scenes the Sobel filter as well as Laplacian of Gaussian also serve as suitable results and are alternatives to global density thresholds. Nevertheless, the improvements with the Canny filter for object edge detection outperform the other techniques, ensuring nearly gapless reconstructions for objects and subsurface in all scenes. 
The trade-off between correctness and completeness with global density thresholds is evident: Lower density thresholds lead to higher completeness but not necessarily superior correctness. The density gradients, especially based on the Canny filter, strike a favorable balance between correctness and completeness across the scenes.

Although our framework achieves high accuracy on the object surfaces, it should be noted that points exist within the objects due to the addressing of the whole voxelized density field. These artifactual points distort the quantitative correctness and do not contribute to the visual appearance and surface accuracy. Limitations are given by processing within the entire density grid thus causing artifacts within the object. This issue may be addressed by extracting only the surface, e.g., using convex hull algorithms. In addition, we aim to apply neural methods for 3D edge detection.

The range and values of the density among different NeRFs, hyperparameters, network configurations, and scenes is variable. Dealing with absolute density values presents a challenge. The density gradients are almost independent of absolute density magnitudes and relying on relative neighboring values in all directions by applying 3D edge detection filter. A notable advantage of our approach is its applicability to different applications. Density gradients allow us to extract surfaces along lower density values using 3D edge detection filters such as Sobel, Canny and Laplacian of Gaussian. 

\section{CONCLUSION}

In summary, we have demonstrated that density gradients based geometric reconstructions lead to high completeness and adequate accuracy.
In considering specific relative density variations or gradients based on first and second derivatives, our approach shows potential for application to various NeRFs, that allow the extraction of a regular voxelized density field. This makes our approach rather independent from absolute NeRFs density output. 
Furthermore, by filtering with global density thresholds, the points are emphasized individually and independently. In contrast, the utilization of 3D gradient filters leverages the inclusion of gradient-based neighborhood information in an anisotropic manner.
Therefore, our approach provides a promising anisotropic solution for complete 3D reconstruction from NeRF with high geometric accuracy.
Consequently, our method introduces a promising, from absolute density magnitude independent solution, which opens new possibilities of reconstructions using NeRFs.

\vspace{1cm}
{
	\begin{spacing}{1.17}
		\normalsize
		\bibliography{ISPRSguidelines_authors} 
	\end{spacing}
}

\end{document}